\documentclass[11pt]{article}

\PassOptionsToPackage{table}{xcolor}
\usepackage[preprint]{acl}

\usepackage{times}
\usepackage{newtxtext}
\usepackage{lmodern}
\usepackage{latexsym}
\usepackage[T1]{fontenc}
\usepackage[utf8]{inputenc}
\usepackage{microtype}
\usepackage{inconsolata}
\usepackage{graphicx}

\usepackage{amsmath}
\usepackage{amsfonts}
\usepackage{amssymb}
\usepackage{amsthm}
\usepackage{algorithm}
\usepackage{algorithmic}
\usepackage{multirow}
\usepackage{booktabs}

\usepackage[most]{tcolorbox}      % 提供 tcolorbox 环境
\usepackage{listings}             % 提供 lstlisting 环境（用于 SQL 高亮）
\usepackage{xcolor}               % 提供颜色定义（\definecolor）
\usepackage{amsmath}              % 提供 equation, aligned 等数学环境
\usepackage{amsfonts}             % 提供 \mathcal, \mathbb 等数学字体
\usepackage{amssymb}              % 提供 \varnothing 等符号
\usepackage{amsthm}

\title{Agent Skills Matter:\\
Inferring Proprietary Skills from Execution Trajectories}

\author{\normalfont
    Jianing Geng\textsuperscript{1}%
    \thanks{This work was done during Jianing Geng's internship at Shanghai Artificial Intelligence Laboratory.}%
    \thanks{Equal contribution.}\quad
    Ruiqi He\textsuperscript{1,5}\footnotemark[2]\quad
  Zekun Fei\textsuperscript{1}\quad
  Biao Yi\textsuperscript{4}\quad
  Xuansheng Wu\textsuperscript{2}\quad
  \\
  Ruijie Wang\textsuperscript{3}\quad
  Zheli Liu\textsuperscript{1}\quad
  Xia Hu\textsuperscript{2}\quad 
  Qingkai Zeng\textsuperscript{1}\thanks{Corresponding author.}\quad
  \\
  \textsuperscript{1}Nankai University\quad
  \textsuperscript{2}Shanghai Artificial Intelligence Laboratory\quad
  \textsuperscript{3}Beihang University\\
  \textsuperscript{4}East China University of Science and Technology\quad 
  \textsuperscript{5}Alibaba Group
  \\
  \texttt{\{gengjianing, heruiqi\}@mail.nankai.edu.cn}\quad
  \texttt{qingkai.zeng@nankai.edu.cn}
}

\begin{document}
\maketitle

% Discourage paragraph endings that leave only one or two words on the last
% line, while retaining enough flexibility for ACL's narrow columns.
\setlength{\parfillskip}{0pt plus 0.5\columnwidth}

\begin{abstract}
Agent skills package reusable procedures that improve downstream performance.
Their lightweight, portable form enables marketplace monetization and private
deployment behind cloud-hosted agent interfaces, giving providers incentives to keep high-value skills proprietary. Yet hiding the artifacts
does not conceal their behavioral effects, which remain observable in
execution trajectories and form a behavioral side channel. We define this
exposure as \textbf{\textit{Skill Leakage}}: reconstructing proprietary skills
from trajectories elicited by benign queries, without reference answers or
success labels. We introduce \textbf{\textsc{SigLeak}}, a black-box framework
that exploits recurring \textit{skill signatures} in agent behavior. It
constructs diverse, decision-rich diagnostic tasks, contrasts matched
skill-enabled and skill-disabled trajectories, and iteratively refines a
reconstructed skill from the isolated patterns. Across five scenarios, three
model families, and three agent frameworks, \textsc{SigLeak} outperforms or
matches three baselines in nearly every setting. It raises success rate by
6.88 percentage points over the skill-disabled reference on average and
achieves the highest overall SkillSim, our metric for coarse- and fine-grained
semantic similarity. These results show that benign execution trajectories can
expose proprietary procedural knowledge. The code is available at \url{https://anonymous.4open.science/r/SigLeak-D1DB}.
\end{abstract}

\section{Introduction}
\label{sec:intro}

% --- skill引出 ---
Large language models (LLMs) are increasingly deployed as autonomous agents, evolving from single-prompt invocations into frameworks capable of multi-step reasoning~\cite{yao2022react,wang2023voyager,yang2024swe,xi2025rise}. To enhance adaptability, developers adopt modular abstractions to structure and extend agent behavior. Among these, \textit{skills}~\cite{AnthropicSKILL,xu2026agent} have emerged as a key mechanism for encapsulating reusable capabilities.

%\textit{Skills} are lightweight, task-specific artifacts dynamically loaded during agent reasoning, combining task instructions, workflow constraints, tool logic, and associated resources~\cite{AnthropicSKILL,xu2026agent}.

\begin{figure}
    \centering
    \includegraphics[width=\linewidth]{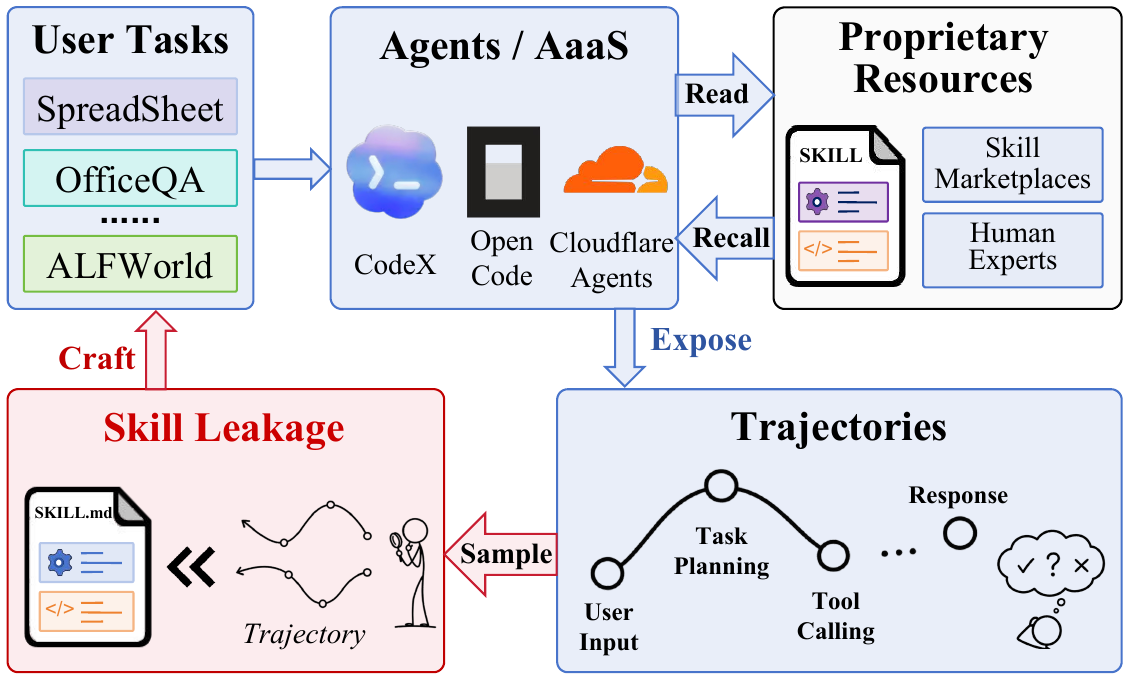}
    \caption{Overview of Skill Leakage. Proprietary skills may leave behavioral signatures in execution trajectories, which an adversary can exploit to infer procedural knowledge without accessing the skill artifact.}
    \label{fig:skill_leakage}
    \vspace{-15pt}
\end{figure}

% === SKILL-important and valuable ===
By distilling domain knowledge into reusable components, skills can substantially improve downstream task efficiency and success rates~\cite{alzubi2026evoskill,yang2026skillopt}. Their lightweight, portable form facilitates reuse across agents and deployment environments while enabling commercial distribution. This commercialization is already visible in paid marketplaces: as of July 2026, Claw Mart reported over \$100,000 in creator payouts~\cite{clawmart2026}. 
%Their lightweight, portable form also facilitates reuse across agents and deployment environments. This portability has fostered paid marketplaces such as ClawMart~\cite{clawmart2026}, where skills can be distributed and monetized. 
Beyond marketplace distribution, emerging Agent-as-a-Service (AaaS) platforms allow providers to deploy skill-equipped agents~\cite{cloudflareagents,awsagentcore}. In such black-box deployments, skills are formalized as core intellectual property (IP) with strictly proprietary attributes. Providers retain these artifacts in controlled backends while exposing only the resulting agent capabilities through service interfaces, thereby preserving both their technical privacy and commercial viability.

% === Real-world harm ===
However, agent interfaces surface tool activity, terminal outputs, and execution progress, allowing users to interrupt or redirect execution~\cite{OpenAICodexCLI,HermesAgent}. Such visibility facilitates oversight and correction, which can improve complex-task performance~\cite{feng-etal-2024-large,zou-etal-2026-llm}. Consequently, backend retention does not conceal a skill's behavioral effects. An adversary may exploit these traces to infer and independently deploy a functional approximation, enabling unauthorized replication or redistribution and undermining providers' intellectual property and competitive advantage. Yet this trajectory-based threat remains underexplored. As illustrated in Figure~\ref{fig:skill_leakage}, we define \textbf{\textit{Skill Leakage}} as the unauthorized inference of proprietary skill content from execution trajectories observed through black-box interactions with the target agent.

% === Distinction from prior work ===
\textit{Skill Leakage} differs from adversarial skill stealing and cooperative skill construction in two respects. \textbf{(1)} Adversarial skill stealing pursues the same objective of recovering hidden skill content but relies on intrusive extraction prompts designed to elicit protected instructions directly~\cite{wang2026black}. Such prompts make the adversary's skill-stealing intent explicit. In contrast, \textit{Skill Leakage} infers hidden skill content indirectly from execution trajectories elicited solely by \emph{benign task queries}. \textbf{(2)} Skill generation methods typically distill reusable procedures from trajectories treated as informative demonstrations, while skill evolution methods optimize accessible skills using explicit success/failure feedback~\cite{zhou2026colleague,ni2026trace2skill,yang2026skillopt}. In \textit{Skill Leakage}, the target skill remains hidden and the observed trajectories are \emph{unlabeled}: neither reference answers nor trajectory-level correctness annotations are available.

% ===== 我们的方法 =====
%making them difficult to distinguish from legitimate task requests

To study this threat, we propose \textbf{\textsc{SigLeak}} (\textbf{Sig}nature-based \textbf{Leak}age Inference of Proprietary Skills), a trajectory-based framework that infers proprietary skills using only \emph{benign task queries}. Its probes never request protected skill content and reveal no overt skill-stealing intent. Our key insight is that skills systematically shape agent execution~\cite{AnthropicSkillCreator}, leaving recurring behavioral cues in observable trajectories. We term these cues \textit{skill signatures}. \textsc{SigLeak} builds on two ideas for eliciting and isolating signatures: \emph{diagnostic task construction} generates diverse, decision-rich probes that expose skill-dependent behavior, while \emph{contrastive trajectory comparison} isolates such behavior by comparing trajectories produced by the same agent for the same probe when instructed to use or not invoke the target skill.

\textsc{SigLeak} realizes these two ideas through a \emph{Generator} and a \emph{Skill Synthesizer}, respectively, within a two-stage workflow. In Stage~1, the Generator uses only the public task description to construct diverse, decision-rich probes, while the Skill Synthesizer compares the resulting trajectory pairs, extracts recurring skill signatures, and consolidates them into an initial reconstruction. In Stage~2, the Generator targets underexplored behavior using the current reconstruction, prior probes, and accumulated signatures, while the Skill Synthesizer incorporates newly extracted signatures. Refinement terminates when no signature warrants an update or the probe budget is exhausted. Neither stage requires reference answers or trajectory-level correctness annotations.

% ===== 实验结果 =====
We evaluate \textsc{SigLeak} against three baselines across five task scenarios, three model families, and three agent frameworks. \textsc{SigLeak} achieves the best or tied-best downstream success rate (SR) in nearly all settings, improving SR over the skill-disabled reference by an average of 6.88 percentage points and attaining the highest SkillSim among inference methods. Additional analyses show that the inferred skills capture target-specific procedural knowledge and that Stage~2 yields its largest gains within the first two refinement rounds.

% ===== 贡献总结 =====
Our contributions are as follows:

\textbf{(1)} We formulate \textbf{\textit{Skill Leakage}} as a trajectory-based threat whereby adversaries infer proprietary skills from unlabeled trajectories elicited by benign diagnostic queries, and provide empirical evidence that distinct functional profiles and trajectory-level \textit{skill signatures} make agent trajectories a behavioral side channel for skill inference.

\textbf{(2)} We propose \textbf{\textsc{SigLeak}}, a two-stage framework that constructs benign diagnostic probes and infers reusable skill instructions through contrastive comparison of paired skill-enabled and skill-disabled trajectories, without access to the original skill or correctness annotations.

\textbf{(3)} We evaluate \textsc{SigLeak} across five task scenarios, three model families, and three agent frameworks, showing that it outperforms inference baselines in downstream utility and semantic fidelity.
    
\begin{figure}
    \centering
    \includegraphics[width=0.95\linewidth]{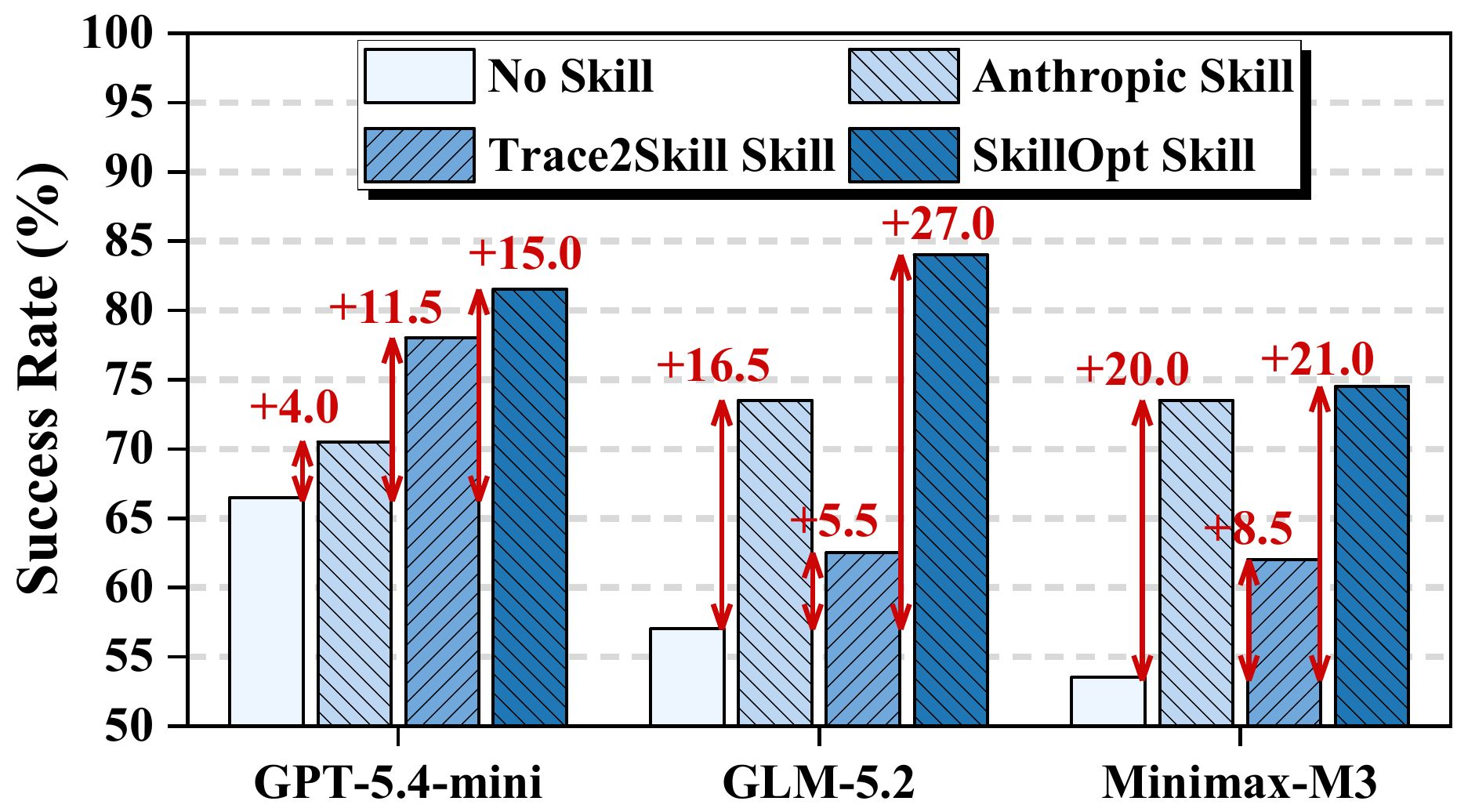}
    \caption{Success rate (\%) on a 100-instance subset of SpreadsheetBench across skill configurations and three model-agent frameworks. See Appendix~\ref{appendix:skillIntro}.}
    \label{fig:skill_fingerprint}
    \vspace{-15pt}

\end{figure}

\section{Preliminaries}
\label{sec:preliminaries}

\paragraph{Definition 1 (Skill-Augmented Agent).}
We consider a tool-augmented LLM agent $\mathcal{A}$ equipped with skill $s$. For a query $q$, the agent yields a user-visible execution trajectory
\begin{equation}
    \tau^{s}
    = \mathcal{A}(q;s)
    = \big[(a_1,o_1),(a_2,o_2),\ldots,(a_T,o_T)\big],
\end{equation}
where $a_t$ denotes an agent action exposed through the interface and $o_t$ denotes the corresponding visible environment observation. Such visibility allows users to monitor, interrupt, or redirect agent execution. We next examine whether execution patterns vary systematically across skills.

\paragraph{Preliminary study.}
To investigate this question, we conduct a preliminary experiment on SpreadsheetBench using three skills with different origins: \textit{Anthropic Skill} (an official skill released by Anthropic), \textit{Trace2Skill}~\cite{ni2026trace2skill} (distilled from large-scale trajectory analysis), and \textit{SkillOpt}~\cite{yang2026skillopt} (iteratively optimized through text-space edits with held-out validation).

Quantitatively, Figure~\ref{fig:skill_fingerprint} reports success rates under
skill configurations across three model-agent frameworks. All skills
outperform the no-skill baseline, but their varying effects across skills and
frameworks reveal distinct functional profiles rather than uniform gains.
SkillOpt achieves the highest performance across all three frameworks,
whereas Trace2Skill yields strong gains on GPT-5.4-mini but smaller gains on
GLM-5.2 and MiniMax-M3. The Anthropic skill provides more consistent
improvements across the two non-GPT frameworks.

Qualitatively, different skills produce distinct execution patterns on the same spreadsheet task, which requires counting table elements and writing the results back to the workbook. The no-skill agent directly reasons about the answer and writes incorrect values. The Anthropic skill exhibits a \textit{formula-first pattern}, generating formulas, attempting recalculation, and repairing cached values. Trace2Skill produces an \textit{explicit-instruction pattern}, preserving the conditional fallback rule and verifying the final cells. SkillOpt produces a \textit{direct-output pattern}, computing values from the workbook and writing literal outputs. These examples show that different skills induce recognizable trajectory-level patterns consistent with their encoded procedural knowledge, rather than merely improving aggregate performance.

\paragraph{Definition 2 (Skill Signatures).}
These results show that skills systematically influence task performance and
execution behavior. For a skill $s$, we define its \emph{skill signatures},
denoted by $\mathcal{S}(s)$, as recurring observable behavioral patterns
through which the procedural knowledge of $s$ manifests in execution
trajectories. They may involve how the agent interprets requirements,
inspects resources, selects tools, recovers from errors, and constructs
outputs.

Given its proprietary value, the provider may deploy $s^*$ in a private backend while withholding its content from users. Nevertheless, its behavioral signatures $\mathcal{S}(s^*)$ remain observable in execution trajectories, creating a trajectory-based leakage channel and posing a practical security risk.

\paragraph{Definition 3 (Skill Leakage Inference).}
We formalize this threat as \emph{Skill Leakage Inference}. We consider a target agent $\mathcal{A}_{s^*}$ equipped with a proprietary skill $s^*$. Given a public task description $d$ and black-box access to the agent, the adversary seeks to infer the procedural content of $s^*$ by constructing
\begin{equation}
    \hat{s}
    =
    \mathcal{G}(d;\mathcal{A}_{s^*}),
\end{equation}
where $\mathcal{G}$ denotes a skill inference method that may rely solely on $d$ or interact with $\mathcal{A}_{s^*}$ through black-box queries. The adversary has no access to the skill document, its associated resources, or the target agent's internal states. No pre-constructed benchmarks, reference answers, or trajectory-level correctness supervision are available.

\paragraph{Motivation.}
The key challenge is attribution: behavior observed in $\tau^s$ may reflect task requirements, base-agent behavior, or model stochasticity rather than the skill itself. This motivates two principles: \emph{diagnostic task construction} uses diverse, decision-rich probes to elicit recurring skill-induced behavior, while \emph{contrastive trajectory comparison} contrasts matched executions with and without the skill to systematically filter out shared behavior.

\begin{figure*}[t]
    \centering
    \includegraphics[width=0.95\linewidth]{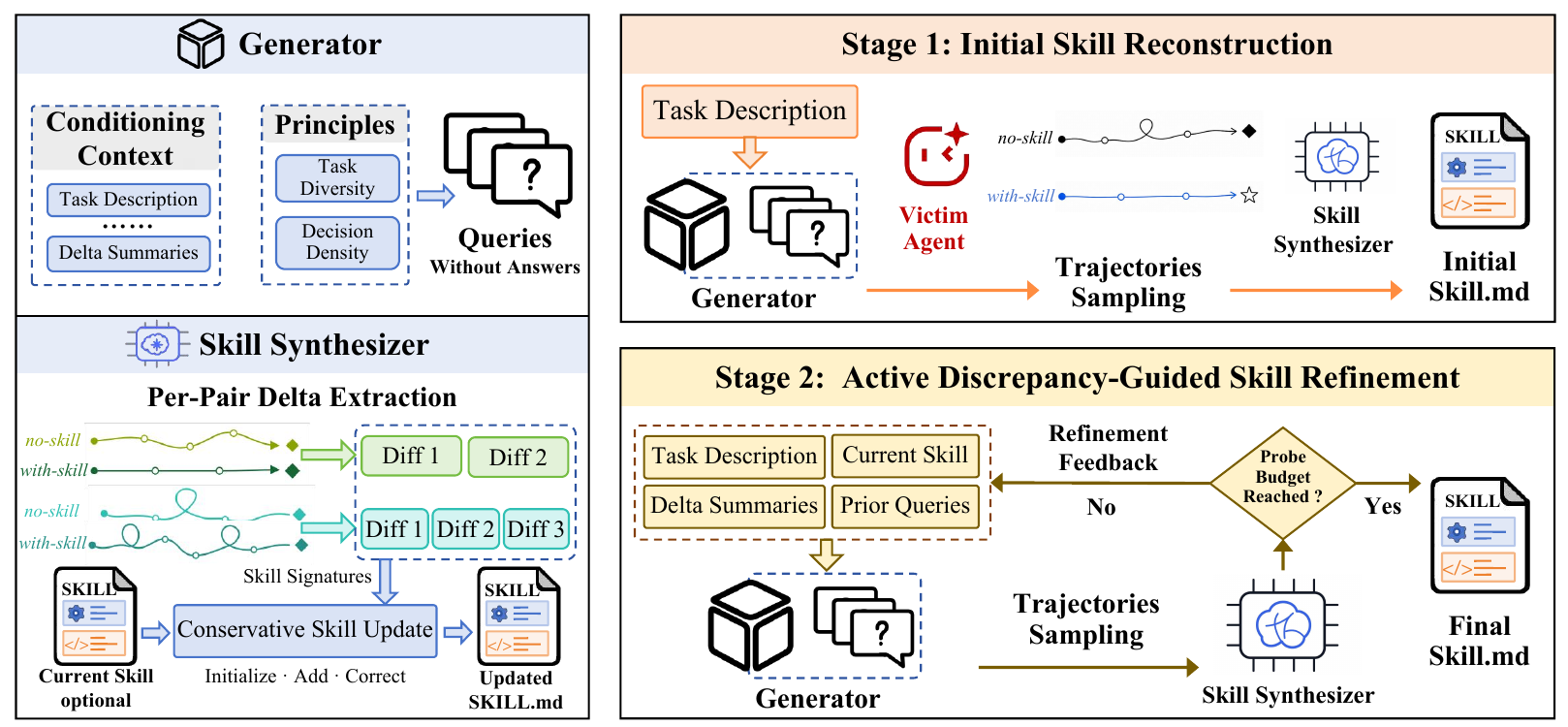}
    \caption{Overview of \textsc{SigLeak}. Stage~1 synthesizes a skill by generating probes and extracting signatures from paired enabled/disabled trajectories. Stage~2 iteratively refines the reconstruction using probes conditioned on the skill, prior probes, and accumulated signatures, until convergence or budget exhaustion. See Appendix~\ref{appendix:methodprompts}.}    \label{fig:overview}
    \vspace{-10pt}

\end{figure*}

\section{Method}
\label{sec:method}

Starting from a public task description $d$, \textsc{SigLeak} reconstructs a hidden proprietary skill through two components. \textbf{(1) \textit{Generator.}} The Generator performs \emph{diagnostic task construction}, producing diverse, decision-rich probes from the conditioning context to expose skill-dependent behavior. \textbf{(2) \textit{Skill Synthesizer.}} The Skill Synthesizer performs \emph{contrastive trajectory comparison}. It contrasts matched trajectories produced for the same probe with and without target-skill invocation, identifies recurring skill signatures, and uses them to initialize or update the reconstructed skill.

Figure~\ref{fig:overview} illustrates the two-stage workflow. \textbf{(1) \textit{Stage~1: Initial skill reconstruction.}} Conditioned solely on $d$, the Generator constructs diagnostic probes, and the Skill Synthesizer consolidates signatures from paired executions into an initial skill $\hat{s}_0$. \textbf{(2) \textit{Stage~2: Discrepancy-guided skill refinement.}} Conditioned on the current skill, prior probes, and accumulated signatures, the Generator targets underexplored behavior, and the Skill Synthesizer updates the skill with new signatures. Refinement continues until convergence or budget exhaustion, yielding the final inferred skill $\hat{s}$.

\subsection{Generator: Diagnostic Task Construction}
\label{sec:generator}
The Generator constructs diagnostic queries to expose skill-dependent behavior. Given context $\mathcal{C}^{(r)}$, it produces a bounded probe set
\begin{equation}
    \mathcal{Q}^{(r)}
    =
    \operatorname{Generate}
    \!\left(\mathcal{C}^{(r)};\Pi\right),
\end{equation}
where $\Pi$ denotes two generation principles: \emph{task diversity} and \emph{decision density}. The stage-specific conditioning contexts are defined separately in Stage~1 and Stage~2.

\emph{Task diversity} encourages probes to span distinct task families rather than constitute superficial paraphrases of the same task. The Generator varies input formats, task constraints, and output requirements, preventing the inferred skill from overfitting to a particular execution pattern. \emph{Decision density} prioritizes probes involving multiple consequential decision points where the hidden skill may influence agent behavior, including requirement interpretation, resource inspection, tool selection, error recovery, and output construction.

\subsection{Skill Synthesizer: Trajectory Comparison}
\label{sec:synthesizer}

\noindent\textbf{Paired execution protocol.}
Existing agent platforms commonly employ progressive disclosure for skill
use: the LLM first receives only skill metadata, such as names and
descriptions, and loads the full skill content only after selecting it for
the current task~\cite{HermesAgent,AnthropicSKILL}. Because skill selection
remains model-mediated, explicit user instructions can affect whether a
skill is invoked.

\textsc{SigLeak} treats the default execution in which the target skill
$s^*$ is invoked as \emph{skill-enabled}. It uses \emph{skill-disabled} as
shorthand for a prompted non-invocation condition: the target skill remains
installed and hidden, but the agent is explicitly instructed not to invoke
it. Specifically, \textsc{SigLeak} prepends a non-invocation instruction
$c^{\mathrm{off}}$ (e.g., ``Do not use any skills.'') without altering the
task content. The complete instruction is provided in
Appendix~\ref{app:skill_disabled_prompt}. For any probe $q$, the paired
prompts are
\begin{equation}
    \tilde{q}^{\mathrm{on}}=q,
    \qquad
    \tilde{q}^{\mathrm{off}}
    =
    c^{\mathrm{off}}\,\Vert\,q,
\end{equation}
where $\Vert$ denotes prompt concatenation. The task component $q$ is held fixed
between conditions, with only the non-invocation instruction differing. The
resulting trajectories are
\begin{equation}
    \tau^{\mathrm{on}}(q)
    =
    \mathcal{A}_{s^*}\!\left(\tilde{q}^{\mathrm{on}}\right),
    \
    \tau^{\mathrm{off}}(q)
    =
    \mathcal{A}_{s^*}\!\left(\tilde{q}^{\mathrm{off}}\right).
\end{equation}

\noindent\textbf{Attacker-visible trajectories.}
Under the skill-inference threat model (Definition~3), the Skill Synthesizer
receives only the user-visible trajectories specified in Definition~1. These
trajectories may include agent messages, tool calls, and environment
observations. The target skill document remains inaccessible to the Skill
Synthesizer.

\noindent\textbf{Signature identification.}
Following Definition~2, the Skill Synthesizer operationalizes skill-signature identification through contrastive trajectory comparison. For the probe set $\mathcal{Q}^{(r)}$, it computes the round-specific set of skill signatures
\begin{equation}
    \mathcal{S}^{(r)}
    =
    \bigcup_{q\in\mathcal{Q}^{(r)}}
    \operatorname{Delta}
    \bigl(q;\tau^{\mathrm{on}}(q),\tau^{\mathrm{off}}(q)\bigr),
\end{equation}
where $\operatorname{Delta}$ is a set-valued operator that maps observable differences in each trajectory pair to candidate task-local signatures, and $\mathcal{S}^{(r)}$ denotes their aggregation in round $r$. Holding the agent and probe fixed controls for shared task requirements and base-agent behavior, making the signatures more attributable to the target skill. We denote signatures accumulated before round $r$ by $\mathcal{S}_{<r}=\bigcup_{k=0}^{r-1}\mathcal{S}^{(k)}$.
%The Skill Synthesizer operates in two modes. In initialization mode, it consolidates recurring cross-task signatures into reusable instructions. In refinement mode, it updates the current inferred skill $\hat{s}_{r-1}$ using newly extracted signatures by adding missing instructions, clarifying underspecified ones, or correcting inconsistent ones.

\subsection{Stage 1: Initial Skill Reconstruction}
\label{sec:initial}
We treat Stage~1 as the initial round $r=0$. Conditioned solely on the task description, i.e., $\mathcal{C}^{(0)}=\{d\}$, the Generator constructs a fixed discovery set $\mathcal{Q}^{(0)}$ comprising six complementary probes. These probes target reasoning, tool selection, intermediate validation, error recovery, output formatting, and task-description expansion. Together, they provide broad initial coverage, while the diversity and decision-density principles avoid behaviorally redundant probes.

Paired execution of $\mathcal{Q}^{(0)}$ yields the initial signature set $\mathcal{S}^{(0)}$. The Skill Synthesizer operates in initialization mode
\begin{equation}
    \hat{s}_0
    =
    \operatorname{Initialize}
    \!\left(\mathcal{S}^{(0)}\right),
\end{equation}
where $\hat{s}_0$ denotes the initial reconstructed skill. This initialization prioritizes coverage by consolidating recurring signatures across diverse tasks into reusable instructions.

\subsection{Stage 2: Discrepancy-Guided Skill Refinement}
\label{sec:refinement}
Stage~2 expands the probe coverage established in Stage~1 and iteratively refines the current reconstruction using newly identified skill signatures. At each refinement round $r\geq1$, the Generator is conditioned on the task description, the current reconstruction $\hat{s}_{r-1}$, summaries of prior probes, and the accumulated skill signatures
\begin{equation}
    \mathcal{C}^{(r)}
    =
    \left\{
        d,\,
        \hat{s}_{r-1},\,
        \operatorname{Summ}(\mathcal{Q}_{<r}),\,
        \mathcal{S}_{<r}
    \right\},
\end{equation}
where $\mathcal{Q}_{<r}=\bigcup_{k=0}^{r-1}\mathcal{Q}^{(k)}$. This context allows the Generator to target underexplored behavior and unresolved discrepancies while avoiding semantic overlap with prior probes.

The new probe set $\mathcal{Q}^{(r)}$ is executed under the same paired protocol, yielding a new signature set $\mathcal{S}^{(r)}$. The Skill Synthesizer uses these signatures to update the current skill
\begin{equation}
    \hat{s}_r
    =
    \operatorname{Update}
    \!\left(\hat{s}_{r-1},\mathcal{S}^{(r)}\right),
\end{equation}
where $\operatorname{Update}$ adds instructions missing from the current reconstruction, elaborates underspecified instructions, and corrects content contradicted by the newly observed signatures. It leaves existing instructions unchanged unless the newly observed signatures indicate that they should be revised.

\paragraph{Termination.}
Refinement terminates if newly extracted signatures yield no update to the current reconstruction, i.e., $\hat{s}_r=\hat{s}_{r-1}$, or if the probe budget $N$ is exhausted. Let $r^*$ denote the termination round. \textsc{SigLeak} returns $\hat{s}=\hat{s}_{r^*}$ as the final reconstructed skill.

\section{Experiments}
\label{sec:evaluation}

\subsection{Experimental Setup}
\label{sec:setup}

\textit{\textbf{Benchmarks.}}
We evaluate \textsc{SigLeak} on five scenarios. \textit{SpreadsheetBench}~\cite{ma2024spreadsheetbench} evaluates spreadsheet manipulation and reasoning. \textit{OfficeQA}~\cite{opsahl2026officeqa} evaluates document-grounded question answering. \textit{SealQA}~\cite{pham2025sealqa} evaluates search-augmented fact seeking under noisy or conflicting web evidence. \textit{LiveMathematicianBench}~\cite{he2026livemathematicianbench} (\textit{LiveMath} in tables) evaluates theorem-grounded reasoning over recent arXiv papers. \textit{ALFWorld}~\cite{shridharalfworld} evaluates sequential decision-making in household environments.

\textit{\textbf{Agents.}}
Our evaluation spans three model families and three agent frameworks. GLM-5.2~\cite{GLM52} and MiniMax-M3~\cite{MiniMaxM3} serve as backends for OpenCode~\cite{OpenCode} and Hermes~\cite{HermesAgent}, while GPT-series models~\cite{GPT54Mini} are evaluated through Codex~\cite{OpenAICodexCLI}.

\textit{\textbf{Baselines.}}
We compare \textsc{SigLeak} against three baselines. \textit{Direct Generation} prompts the inference model to synthesize a skill directly from the task description. \textit{Naive Trace Summarization} uses the Generator to construct probes, collects only skill-enabled trajectories, and summarizes them into a skill. 
\textit{BBS}~\cite{wang2026black} is a prompt-stealing baseline that uses explicitly adversarial extraction instructions. To model a defended deployment, we equip every target agent with the SkillGuard5 prompt-defense instruction introduced by BBS, regardless of the inference method. BBS applies its 12 categories of prompt-stealing queries to recover the target skill. See Appendix~\ref{appendix:baseline} for implementation details.

The same task description is provided to all compared inference methods, with
the descriptions included in
Appendix~\ref{appendix:experimental_configuration}.

\textit{\textbf{SigLeak configuration.}}
For each scenario, \textsc{SigLeak} receives only a brief task description of
the setting in which the target skill is intended to operate. The same task
description is provided to all compared inference methods, with the descriptions included in Appendix~\ref{appendix:experimental_configuration}. Stage~1 generates six
diagnostic probes. Stage~2 generates three probes per round and runs for at
most two rounds in the main experiments. We evaluate methods under their
native interaction protocols: Direct Generation is non-interactive, Naive
uses six skill-enabled executions, BBS uses 12 extraction queries, and
\textsc{SigLeak}'s 12 paired probes. We use DeepSeek-V4-Flash~\cite{DeepSeekV4Flash} for the Generator and Skill
Synthesizer across all configurations.

\subsection{Evaluation Metrics}
\label{sec:metrics}
\textit{\textbf{Downstream Success Rate (SR).}}
Success is measured using the official or task-specific evaluator for each benchmark: spreadsheet correctness for SpreadsheetBench, answer accuracy for OfficeQA and SealQA, mathematical correctness for LiveMathematicianBench, and task completion for ALFWorld. Because long-horizon agent executions are stochastic, we evaluate each configuration over three independent runs and report the mean success rate, $\overline{\mathrm{SR}}=\frac{1}{3}\sum_{j=1}^{3}\mathrm{SR}_j$, where $\mathrm{SR}_j$ denotes the success rate of run $j$. We report mean SR with the skill disabled, the target skill $s^*$ enabled, and the inferred skill $\hat{s}$ enabled. Higher SR under $\hat{s}$ indicates greater recovery of the target skill's functional benefit.

\textit{\textbf{Skill Similarity (SkillSim).}}
We evaluate semantic inference fidelity at coarse and fine-grained levels.
\emph{Coarse SkillSim} uses an LLM judge to score the overall behavioral
similarity between the reference and inferred skills on a five-point scale.
\emph{Fine SkillSim} extracts source-grounded Skill Behavior Constraints
(SBCs)~\cite{tan2026skillcoveragetestadequacy} and performs graded one-to-one
matching between reference and inferred SBCs. Let $S$ denote the total weighted
match mass, and let $N_{\mathrm{inf}}$ and $N_{\mathrm{ref}}$ denote the numbers
of inferred and reference SBCs, respectively. We compute
\begin{equation}
\begin{aligned}
P=\frac{S}{N_{\mathrm{inf}}},
R=\frac{S}{N_{\mathrm{ref}}},
F_1=\frac{2S}{N_{\mathrm{inf}}+N_{\mathrm{ref}}},
\end{aligned}
\end{equation}
where $P$, $R$, and $F_1$ denote soft precision, recall, and F1, reported as percentages. See Appendix~\ref{appendix:skillsim} for details.

\newcommand{\modelagent}[2]{%
  \shortstack[c]{#1\\[-1pt]#2}%
}

\newcommand{\boundpair}[2]{%
  \cellcolor{gray!12}%
  \raisebox{-0.55ex}{#1}\hspace{-0.10em}%
  \raisebox{0.00ex}{\rotatebox{42}{\rule{0.35pt}{1.05em}}}%
  \hspace{-0.20em}\raisebox{0.55ex}{#2}%
}
\newcommand{\boundlabel}{%
  Skill Bounds~(\raisebox{-0.55ex}{w/o}\hspace{-0.10em}%
  \raisebox{0.00ex}{\rotatebox{42}{\rule{0.35pt}{1.05em}}}%
  \hspace{-0.20em}\raisebox{0.55ex}{w/})%
}
\newcommand{\shade}[1]{\cellcolor{gray!6}#1}
\newcommand{\headcell}[1]{{\bfseries\shortstack[c]{#1}}}

% Upper script: delta versus no-skill. Lower script: delta versus with-skill.
% Both scripts share the same horizontal origin immediately after the value.
\newcommand{\deltamarks}[2]{%
  \makebox[0pt][l]{\raisebox{0.90ex}{\scalebox{0.78}{$\scriptscriptstyle #1$}}}%
  \makebox[0pt][l]{\raisebox{-0.50ex}{\scalebox{0.78}{$\scriptscriptstyle #2$}}}%
}
\newcommand{\entry}[3]{\makebox[4.35em][l]{#1\deltamarks{#2}{#3}}}
\newcommand{\bestentry}[3]{\makebox[4.35em][l]{\textbf{#1}\deltamarks{#2}{#3}}}

\begin{table*}[!t]
\centering
\small
\setlength{\tabcolsep}{1.85pt}
\renewcommand{\arraystretch}{1.00}

\begin{tabular}{llccccccc}
\toprule
\multirow{2}{*}{\headcell{Scenario}}
& \multirow{2}{*}{\headcell{Setting / Method}}
& \headcell{GPT-5.4}
& \headcell{GPT-5.5}
& \headcell{GLM-5.2}
& \headcell{MM-M3}
& \headcell{GLM-5.2}
& \headcell{MM-M3}
& \multirow{2}{*}{\headcell{Avg}} \\
& & \headcell{Codex} & \headcell{Codex}
& \headcell{OC} & \headcell{OC}
& \headcell{HM} & \headcell{HM} & \\
\midrule

\multirow{5}{*}{\textbf{Spreadsheet}}
& \cellcolor{gray!12}\textit{\boundlabel}
& \boundpair{69.33}{79.17} & \boundpair{79.83}{86.17} & \boundpair{72.33}{92.67}
& \boundpair{54.33}{85.67} & \boundpair{79.83}{84.67} & \boundpair{73.33}{92.67} & \boundpair{71.50}{86.84} \\
\cmidrule(lr){2-9}
& Direct Generation
& \entry{68.67}{-0.66}{-10.50} & \entry{80.17}{+0.34}{-6.00} & \entry{83.83}{+11.50}{-8.84}
& \entry{77.67}{+23.34}{-8.00} & \entry{79.83}{+0.00}{-4.84} & \entry{81.17}{+7.84}{-11.50} & \entry{78.56}{+7.06}{-8.28} \\
& Naive Trace
& \shade{\entry{72.00}{+2.67}{-7.17}} & \shade{\entry{80.50}{+0.67}{-5.67}} & \shade{\entry{86.17}{+13.84}{-6.50}}
& \shade{\entry{79.83}{+25.50}{-5.84}} & \shade{\entry{81.17}{+1.34}{-3.50}} & \shade{\entry{84.83}{+11.50}{-7.84}} & \shade{\entry{80.75}{+9.25}{-6.09}} \\
& BBS
& \entry{69.33}{+0.00}{-9.84} & \entry{79.83}{+0.00}{-6.34} & \entry{72.33}{+0.00}{-20.34}
& \entry{54.33}{+0.00}{-31.34} & \entry{79.83}{+0.00}{-4.84} & \entry{73.33}{+0.00}{-19.34} & \entry{71.50}{+0.00}{-15.34} \\
& \textbf{SigLeak}
& \shade{\bestentry{76.17}{+6.84}{-3.00}} & \shade{\bestentry{83.17}{+3.34}{-3.00}} & \shade{\bestentry{90.67}{+18.34}{-2.00}}
& \shade{\bestentry{83.67}{+29.34}{-2.00}} & \shade{\bestentry{83.67}{+3.84}{-1.00}} & \shade{\bestentry{90.17}{+16.84}{-2.50}} & \shade{\bestentry{84.59}{+13.09}{-2.25}} \\

\addlinespace[2pt]
\midrule
\multirow{5}{*}{\textbf{OfficeQA}}
& \cellcolor{gray!12}\textit{\boundlabel}
& \boundpair{61.65}{68.42} & \boundpair{77.44}{84.21} & \boundpair{64.66}{75.19}
& \boundpair{69.17}{75.94} & \boundpair{64.91}{76.69} & \boundpair{39.10}{51.13} & \boundpair{62.82}{71.93} \\
\cmidrule(lr){2-9}
& Direct Generation
& \entry{64.41}{+2.76}{-4.01} & \entry{77.44}{+0.00}{-6.77} & \entry{68.67}{+4.01}{-6.52}
& \entry{70.68}{+1.51}{-5.26} & \entry{76.19}{+11.28}{-0.50} & \entry{44.36}{+5.26}{-6.77} & \entry{66.96}{+4.14}{-4.97} \\
& Naive Trace
& \shade{\entry{63.91}{+2.26}{-4.51}} & \shade{\entry{78.20}{+0.76}{-6.01}} & \shade{\entry{67.17}{+2.51}{-8.02}}
& \shade{\entry{68.92}{-0.25}{-7.02}} & \shade{\entry{76.19}{+11.28}{-0.50}} & \shade{\entry{43.36}{+4.26}{-7.77}} & \shade{\entry{66.29}{+3.47}{-5.64}} \\
& BBS
& \entry{61.65}{+0.00}{-6.77} & \entry{77.44}{+0.00}{-6.77} & \entry{64.66}{+0.00}{-10.53}
& \entry{69.17}{+0.00}{-6.77} & \entry{64.91}{+0.00}{-11.78} & \entry{39.10}{+0.00}{-12.03} & \entry{62.82}{+0.00}{-9.11} \\
& \textbf{SigLeak}
& \shade{\bestentry{66.92}{+5.27}{-1.50}} & \shade{\bestentry{79.45}{+2.01}{-4.76}} & \shade{\bestentry{69.42}{+4.76}{-5.77}}
& \shade{\bestentry{73.93}{+4.76}{-2.01}} & \shade{\bestentry{76.94}{+12.03}{+0.25}} & \shade{\bestentry{49.12}{+10.02}{-2.01}} & \shade{\bestentry{69.30}{+6.48}{-2.63}} \\

\addlinespace[2pt]
\midrule
\multirow{5}{*}{\textbf{SealQA}}
& \cellcolor{gray!12}\textit{\boundlabel}
& \boundpair{42.04}{51.35} & \boundpair{53.75}{70.27} & \boundpair{44.44}{50.45}
& \boundpair{44.14}{52.25} & \boundpair{41.44}{47.75} & \boundpair{34.53}{46.85} & \boundpair{43.39}{53.15} \\
\cmidrule(lr){2-9}
& Direct Generation
& \entry{44.14}{+2.10}{-7.21} & \entry{62.76}{+9.01}{-7.51} & \entry{45.05}{+0.61}{-5.40}
& \entry{47.75}{+3.61}{-4.50} & \entry{43.24}{+1.80}{-4.51} & \entry{34.83}{+0.30}{-12.02} & \entry{46.29}{+2.90}{-6.86} \\
& Naive Trace
& \shade{\entry{42.64}{+0.60}{-8.71}} & \shade{\entry{60.66}{+6.91}{-9.61}} & \shade{\entry{44.74}{+0.30}{-5.71}}
& \shade{\entry{49.55}{+5.41}{-2.70}} & \shade{\entry{42.34}{+0.90}{-5.41}} & \shade{\entry{33.33}{-1.20}{-13.52}} & \shade{\entry{45.54}{+2.15}{-7.61}} \\
& BBS
& \entry{42.04}{+0.00}{-9.31} & \entry{53.75}{+0.00}{-16.52} & \entry{44.44}{+0.00}{-6.01}
& \entry{44.14}{+0.00}{-8.11} & \entry{41.44}{+0.00}{-6.31} & \entry{34.53}{+0.00}{-12.32} & \entry{43.39}{+0.00}{-9.76} \\
& \textbf{SigLeak}
& \shade{\bestentry{49.85}{+7.81}{-1.50}} & \shade{\bestentry{66.97}{+13.22}{-3.30}} & \shade{\bestentry{47.75}{+3.31}{-2.70}}
& \shade{\bestentry{52.25}{+8.11}{+0.00}} & \shade{\bestentry{46.85}{+5.41}{-0.90}} & \shade{\bestentry{41.44}{+6.91}{-5.41}} & \shade{\bestentry{50.85}{+7.46}{-2.30}} \\

\addlinespace[2pt]
\midrule
\multirow{5}{*}{\textbf{LiveMath}}
& \cellcolor{gray!12}\textit{\boundlabel}
& \boundpair{24.29}{29.19} & \boundpair{44.63}{55.74} & \boundpair{22.03}{27.12}
& \boundpair{23.35}{27.12} & \boundpair{32.77}{43.50} & \boundpair{28.81}{36.16} & \boundpair{29.31}{36.47} \\
\cmidrule(lr){2-9}
& Direct Generation
& \entry{25.05}{+0.76}{-4.14} & \entry{44.82}{+0.19}{-10.92} & \entry{25.42}{+3.39}{-1.70}
& \entry{28.63}{+5.28}{+1.51} & \entry{36.72}{+3.95}{-6.78} & \entry{26.74}{-2.07}{-9.42} & \entry{31.23}{+1.92}{-5.24} \\
& Naive Trace
& \shade{\entry{23.73}{-0.56}{-5.46}} & \shade{\entry{48.40}{+3.77}{-7.34}} & \shade{\entry{25.05}{+3.02}{-2.07}}
& \shade{\entry{32.77}{+9.42}{+5.65}} & \shade{\entry{27.87}{-4.90}{-15.63}} & \shade{\entry{29.57}{+0.76}{-6.59}} & \shade{\entry{31.23}{+1.92}{-5.24}} \\
& BBS
& \entry{24.29}{+0.00}{-4.90} & \entry{44.63}{+0.00}{-11.11} & \entry{22.03}{+0.00}{-5.09}
& \entry{23.35}{+0.00}{-3.77} & \entry{32.77}{+0.00}{-10.73} & \entry{28.81}{+0.00}{-7.35} & \entry{29.31}{+0.00}{-7.16} \\
& \textbf{SigLeak}
& \shade{\bestentry{27.31}{+3.02}{-1.88}} & \shade{\bestentry{50.09}{+5.46}{-5.65}} & \shade{\bestentry{26.18}{+4.15}{-0.94}}
& \shade{\bestentry{34.09}{+10.74}{+6.97}} & \shade{\bestentry{38.61}{+5.84}{-4.89}} & \shade{\bestentry{34.09}{+5.28}{-2.07}} & \shade{\bestentry{35.06}{+5.75}{-1.41}} \\

\addlinespace[2pt]
\midrule
\multirow{5}{*}{\textbf{ALFWorld}}
& \cellcolor{gray!12}\textit{\boundlabel}
& \boundpair{76.67}{83.57} & \boundpair{95.95}{96.90} & \boundpair{99.29}{100.00}
& \boundpair{91.43}{99.76} & \boundpair{92.14}{92.86} & \boundpair{95.71}{99.76} & \boundpair{91.87}{95.48} \\
\cmidrule(lr){2-9}
& Direct Generation
& \entry{75.48}{-1.19}{-8.09} & \entry{96.19}{+0.24}{-0.71} & \entry{99.52}{+0.23}{-0.48}
& \entry{92.62}{+1.19}{-7.14} & \entry{91.90}{-0.24}{-0.96} & \entry{96.90}{+1.19}{-2.86} & \entry{92.10}{+0.23}{-3.38} \\
& Naive Trace
& \shade{\entry{78.81}{+2.14}{-4.76}} & \shade{\entry{96.90}{+0.95}{+0.00}} & \shade{\bestentry{99.76}{+0.47}{-0.24}}
& \shade{\bestentry{97.62}{+6.19}{-2.14}} & \shade{\bestentry{92.62}{+0.48}{-0.24}} & \shade{\entry{96.90}{+1.19}{-2.86}} & \shade{\bestentry{93.77}{+1.90}{-1.71}} \\
& BBS
& \entry{76.67}{+0.00}{-6.90} & \entry{95.95}{+0.00}{-0.95} & \entry{99.29}{+0.00}{-0.71}
& \entry{91.43}{+0.00}{-8.33} & \entry{92.14}{+0.00}{-0.72} & \entry{95.71}{+0.00}{-4.05} & \entry{91.87}{+0.00}{-3.61} \\
& \textbf{SigLeak}
& \shade{\bestentry{79.52}{+2.85}{-4.05}} & \shade{\bestentry{97.38}{+1.43}{+0.48}} & \shade{\bestentry{99.76}{+0.47}{-0.24}}
& \shade{\entry{94.05}{+2.62}{-5.71}} & \shade{\bestentry{92.62}{+0.48}{-0.24}} & \shade{\bestentry{97.62}{+1.91}{-2.14}} & \shade{\entry{93.49}{+1.62}{-1.99}} \\
\bottomrule
\end{tabular}

\caption{Main skill-inference performance across model-agent configurations
and scenarios. Skill Bounds cells show reference SRs with the target skill
disabled (w/o) and enabled (w/) in the lower-left and upper-right,
respectively. Method entries report mean SR (\%); superscripts/subscripts show
SR differences from the w/o/w/ references in percentage points. Avg averages
across model-agent configurations. OC, HM, and MM-M3 denote OpenCode, Hermes,
and MiniMax-M3, respectively. Bold indicates the best or tied-best inference
result among inference methods.}
\label{tab:main_results_delta}
\end{table*}

\begin{figure*}[t]
    \centering
    \includegraphics[width=\linewidth]{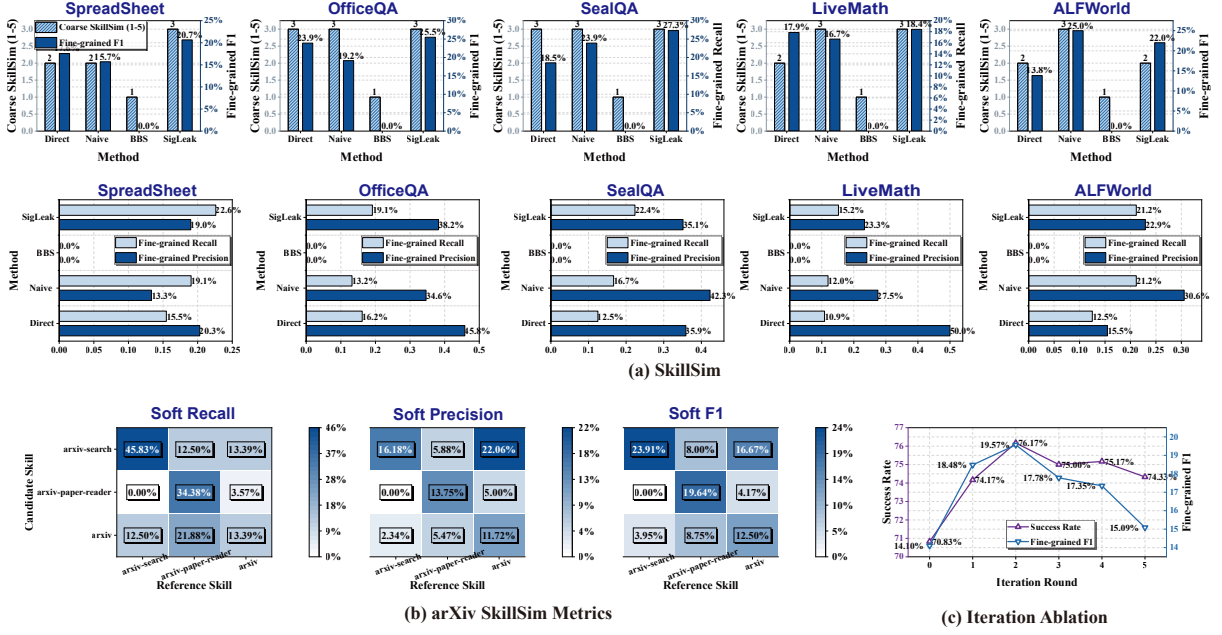}
    \caption{(a) SkillSim results across five scenarios. The top row reports Coarse SkillSim and fine-grained F1, while the bottom row reports fine-grained recall and precision. (b) Pairwise soft recall, precision, and F1 among three arXiv-search skills, testing whether each inferred skill is most similar to its target. (c) Iteration ablation of downstream success rate and fine-grained F1 across the initial inference at round 0 and five refinement rounds.}    \label{fig:skillsim}
    \vspace{-10pt}

\end{figure*}

\subsection{Functional Utility of Inferred Skills}
\label{sec:main_results}

\textit{\textbf{Effectiveness across model-agent configurations.}}
As reported in Table~\ref{tab:main_results_delta}, \textsc{SigLeak} achieves the best or tied-best SR among inference methods across most model-agent combinations. On Codex, it ranks best or tied-best across all GPT-5.4-mini and GPT-5.5 scenarios, with average gains of 5.16 and 5.09 percentage points (pp), respectively. On OpenCode and Hermes, \textsc{SigLeak} yields average gains of 11.11 and 8.19 pp for MiniMax-M3, compared with 6.21 and 5.52 pp for GLM-5.2.

\textit{\textbf{Effectiveness across task scenarios.}}
\textsc{SigLeak} improves average SR over the skill-disabled reference across all five scenarios. Scenario-level improvements range from +1.62 pp on ALFWorld to
+13.09 pp on Spreadsheet, with substantial gains on SealQA (+7.46 pp), OfficeQA (+6.48 pp), and LiveMath (+5.75 pp). This consistent improvement across spreadsheet manipulation, document and web reasoning, mathematical reasoning, and interactive decision-making demonstrates \textsc{SigLeak}'s applicability to heterogeneous skill domains.

\textit{\textbf{Comparison with inference baselines.}}
Aggregated over all model-agent/scenario combinations, \textsc{SigLeak} improves over the skill-disabled reference by 6.88 pp on average. Direct Generation and Naive Trace Summarization also often improve over the skill-disabled reference, but with smaller average gains of 3.25 and 3.74 pp, respectively. With the SkillGuard5 prompt defense, BBS matches the skill-disabled reference, indicating that it recovers no usable skill content. Overall, paired trajectory comparison provides a more robust signal, although the heterogeneous gains suggest sensitivity to trajectory detail and skill observability.

\textit{\textbf{Trajectory-exposed skill signatures.}} 
ALFWorld is an exception, where Naive Trace Summarization is comparable or occasionally superior. Its target skill exposes sequential-control signatures, including state tracking, search bookkeeping, and action ordering, directly in skill-enabled trajectories. This visibility makes direct summarization competitive, while high skill-disabled performance limits further gains. \textsc{SigLeak} is more advantageous when signatures must be disentangled from task structure and base-agent behavior.

\subsection{Semantic Fidelity of Inferred Skills}
\label{sec:fidelity}

\textbf{\textit{\textsc{SigLeak} captures more target-skill content.}}
Figure~\ref{fig:skillsim}(a) reports inference fidelity across five scenarios on GPT-5.4-mini. \textsc{SigLeak} achieves a top Coarse SkillSim score in four scenarios. While Coarse SkillSim summarizes overall similarity, the fine-grained metrics provide a more discriminative comparison. \textsc{SigLeak} ranks first or ties for first in F1 in four scenarios and recall in all five, with the highest average F1 and recall at 22.8\% and 20.1\%, respectively. Naive Trace Summarization achieves 20.1\% F1 and 16.4\% recall, while Direct Generation achieves 18.3\% F1 and 13.5\% recall.

Although Direct Generation and Naive Trace Summarization attain higher average precision at 33.5\% and 29.7\%, this advantage comes with lower recall, indicating recovery of a smaller subset of target constraints. ALFWorld is the only F1 exception, where Naive Trace Summarization slightly outperforms \textsc{SigLeak}. This likely reflects the target skill's directly observable sequential-control signatures, which favor trace summarization. Under the prompt defense, BBS attains zero fine-grained recall and F1, consistent with its downstream performance at the skill-disabled level.

\textbf{\textit{\textsc{SigLeak} distinguishes skills within the same domain.}}
The preceding results establish that \textsc{SigLeak} recovers target-skill content across diverse domains. We next ask a stricter question: \textit{can it distinguish multiple skills that address similar tasks but prescribe different procedures?} To isolate this capability, we conduct a controlled analysis using
three arXiv-related skills that share a common task context but implement distinct workflows. \textsc{SigLeak} infers each skill from the
same task description, so any distinction must arise from its observed trajectory signatures. Figure~\ref{fig:skillsim}(b) reports pairwise soft F1
between inferred skills (rows) and target skills (columns). The resulting matrix is diagonally dominant in every row. For example, the inferred source-level paper-reading skill achieves 19.64\% F1 against its corresponding target, compared with at most 4.17\% against either alternative. These results show that \textsc{SigLeak} recovers target-specific procedural knowledge rather than merely generating a generic prompt for arXiv-related tasks.

\subsection{Effectiveness of Two-Stage Inference}
\label{sec:ablation}

\textbf{\textit{Stage~1 produces an effective initial skill.}}
Before refinement (round 0), the initial skill achieves 70.83\% SR and 14.10\% fine-grained SkillSim F1. This indicates that diagnostic probing and paired trajectory comparison can recover useful functional and semantic content when conditioned solely on the task description.

\textbf{\textit{Stage~2 gains are concentrated in early rounds.}}
Figure~\ref{fig:skillsim}(c) tracks downstream SR and fine-grained SkillSim F1 from round 0 through five rounds. Both peak at round 2, with SR rising from 70.83\% to 76.17\% and F1 from 14.10\% to 19.57\%. Later rounds provide no sustained improvement, with both declining from their peaks. This pattern suggests that early rounds identify actionable signature gaps, whereas later probes yield diminishing or noisier signals, motivating bounded refinement and early termination.

% \subsection{Case Study: \textsc{SigLeak} in Action}
% \label{sec:case_study}

% We examine representative spreadsheet probes from both stages to illustrate how \textsc{SigLeak} progresses from broad signature discovery to targeted refinement. In Stage~1, a cross-sheet editing probe reveals distinct inspection and validation behaviors. Although both executions complete the requested edit, only the skill-enabled agent inspects merged ranges and table definitions before modification and reopens the saved workbook to verify its structure, formulas, and unchanged cells. Together with recurring evidence from other probes, these differences yield an initial workflow.

% Stage~2 targets a remaining gap: the initial skill requires preserving merged layouts but does not specify how to handle them during full-row rewriting. A targeted row-manipulation probe reveals additional rules for auditing styled and formula cells, recording merged ranges before clearing rows, and restoring them afterward. The resulting update turns a broad preservation principle into an executable, operation-specific procedure. Appendix~\ref{appendix:case_study} provides the complete probes, trajectory differences, and skill updates.

\section{Related Work}
\label{sec:related_work}

As LLM agents become increasingly capable, skills have emerged as a practical abstraction for packaging reusable procedures, tool-use conventions, and task-specific resources~\cite{jiang2026sokagenticskills,ding2026agentskillevaluationevolution}. Recent work has investigated their construction, collection, optimization, and distribution, as well as associated security risks.

\textbf{\textit{The Emergence of Agent Skills.}}
Agent skills improve performance by externalizing useful procedures as reusable artifacts~\cite{AnthropicSKILL,xu2026agent}, while recent benchmarks evaluate their effectiveness across diverse tasks~\cite{li2026skillsbench,zhou2026skillgenbenchbenchmarkingskillgeneration,chen2026skillcraftllmagentslearn}. Following Anthropic's Skill Creator~\cite{AnthropicSkillCreator}, recent studies have explored how to construct effective skills for LLM agents~\cite{ma2026skillgen,wang2026skills,alzubi2026evoskill,li2026mind,zhou2026colleague}. Other work collects skills at scale from project repositories and documentation~\cite{liang2026skillnet,karaaslan2026skillseekers,shen2026skillfoundry} or organizes them into reusable libraries~\cite{yang2026autoskill,wang2026skillx,zhou2026memento}. While these studies establish skills as reusable components, \textsc{SigLeak} asks whether their procedural knowledge can be inferred from the execution behavior they induce.

\textbf{\textit{Skill Evolution from Execution Feedback.}}
Another line of work optimizes skills using execution feedback. Some methods analyze execution trajectories to propose new skills or edit existing ones~\cite{ni2026trace2skill,yu2026skilladaptor}, while others continuously refine editable skill artifacts through deployment feedback~\cite{yang2026skillopt,zhang2026skillevolver}. Further studies train or adapt skills for self-evolution~\cite{wang-etal-2026-reinforcement,xia2026skillrl,huang2026skill}. These studies demonstrate that trajectories contain rich behavioral information. In contrast, \textsc{SigLeak} treats trajectories as a leakage channel and infers hidden skill-induced rules from unlabeled paired executions without success/failure supervision.

\textbf{\textit{Skill Ecosystems and Security Implications.}}
Because skills are lightweight and portable~\cite{saha2026skilldex,bhardwaj2026formal}, they have given rise to marketplaces where users can share, buy, or sell reusable capabilities~\cite{skillsmp,claudemarketplaces,clawmart2026,skillsllm2026}. This ecosystem makes proprietary skills valuable assets whose procedural knowledge warrants protection. Even when the underlying artifacts remain hidden, observable executions may expose the knowledge they encode.

\textbf{\textit{Instruction and Trajectory Leakage.}}
Prior work has examined the extraction of protected information from LLM applications, including system prompts~\cite{zhang2024effective, PLeak}, private interaction records stored in agent memory~\cite{wang-etal-2025-unveiling-privacy}, and proprietary agent skills~\cite{wang2026black}. These attacks rely on overtly adversarial prompts designed to induce direct disclosure. RedAct studies procedural leakage from a defender-released corpus of successful traces, with downstream methods operating only on these pre-collected artifacts~\cite{xu2026redact}. In contrast, \textsc{SigLeak} considers an interactive black-box setting. Given only a public task description, the adversary adaptively issues benign probes and contrasts matched executions without outcome labels to isolate skill signatures, requiring no benchmark instances, reference answers, or correctness supervision. To our knowledge, \textsc{SigLeak} is the first framework to formulate and operationalize active skill leakage from benign, unlabeled agent interactions through diagnostic probing and contrastive trajectory comparison.
\section{Conclusion}

We formulate \textit{Skill Leakage}, a trajectory-based threat in which proprietary skill content is inferred from unlabeled, user-visible trajectories elicited through benign diagnostic queries. We propose \textsc{SigLeak}, which combines diagnostic task construction with contrastive analysis of matched skill-enabled and skill-disabled trajectories to isolate skill signatures and infer reusable instructions and workflows. Across five scenarios, three model families, and three agent frameworks, \textsc{SigLeak} achieves the best or tied-best downstream SR in nearly all settings, improves over skill-disabled agents by 6.88 percentage points on average, and attains the highest overall SkillSim among inference methods. Further analyses indicate target-specific procedural recovery. These findings show that proprietary skills remain behaviorally exposed despite artifact concealment and defenses against direct extraction, motivating defenses that mitigate behavioral inference while preserving execution visibility for human oversight.

\section*{Ethical Statement}
Although \textsc{SigLeak} is an attack framework, all experiments were conducted in locally controlled environments using benchmark tasks and experimental agent configurations. We did not probe any deployed commercial agent services or access private user data or nonpublic skill artifacts. This study aims to characterize skill-leakage risks and inform the development of effective defenses.

\section*{Limitations}
Our implementation uses fixed Generator and Skill Synthesizer prompts and a
common probe budget across scenarios. We cap Stage~2 at two rounds based on the
ablation, where both downstream SR and fine-grained F1 peak, but a single
global configuration may not fully exploit scenario-specific trajectory
evidence. Future work could dynamically adapt the probe budget and synthesis
strategy to each scenario.

\bibliography{Bibliography-File}

\appendix
\begin{figure*}[t]
    \centering
    \includegraphics[width=\linewidth]{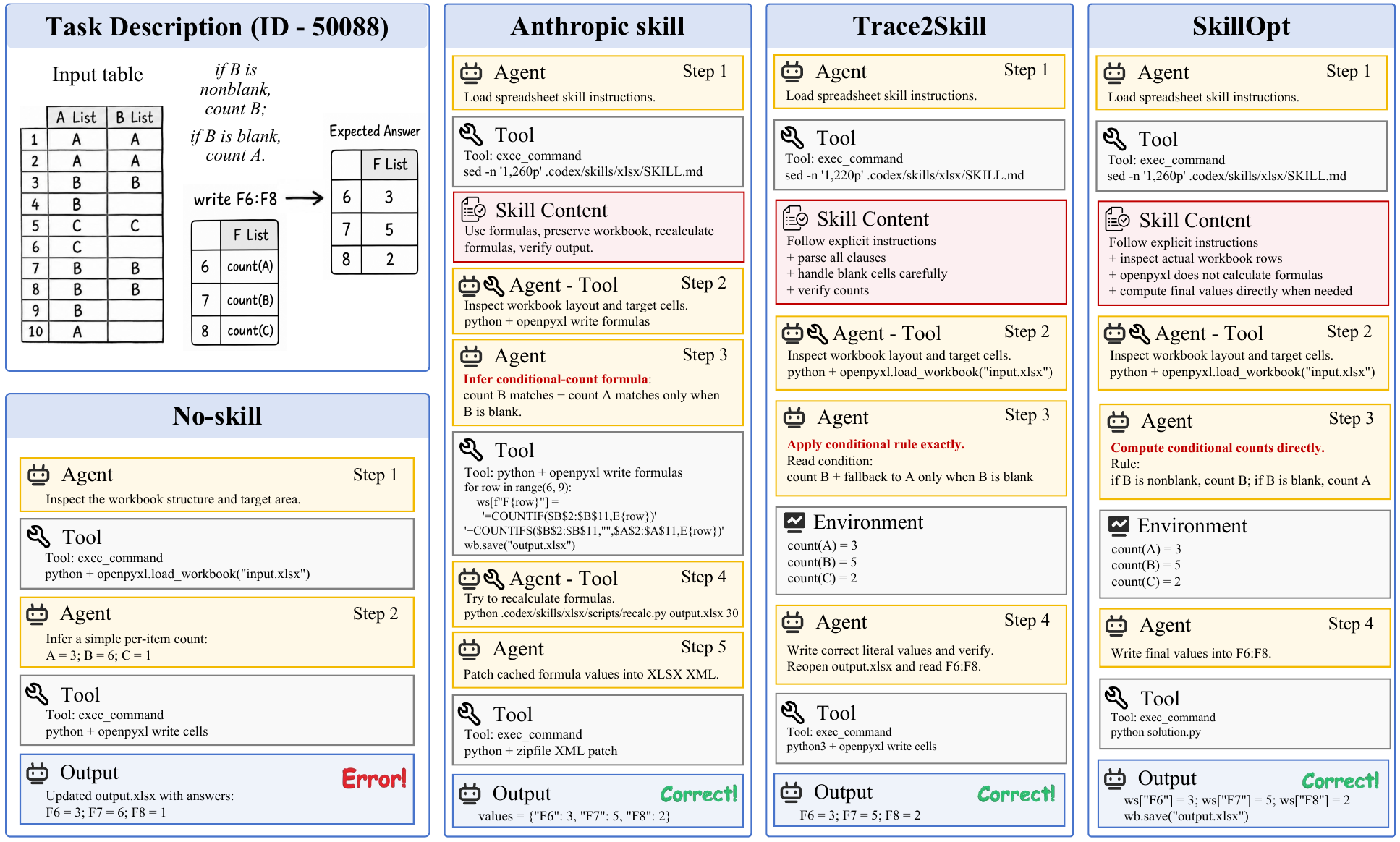}
    \caption{A SpreadsheetBench example (ID 50088). The no-skill agent misses the fallback rule and outputs incorrect values \(3,6,1\). All three skill-enabled agents succeed but exhibit distinct signatures: Anthropic Skill uses formulas, recalculation, and cached-value repair; Trace2Skill emphasizes rule following and post-edit verification; SkillOpt writes literal values. This researcher-side visualization includes skill-loading information only for interpretation and is not used as input to \textsc{SigLeak}.}
    \label{fig:task50088_trace}
\end{figure*}

\section{Preliminary Study}
\label{appendix:skillIntro}

\paragraph{Scope of the preliminary study.}
This section presents a researcher-side qualitative analysis of known skill
configurations and is separate from the \textsc{SigLeak} inference pipeline.
The figures may include skill-loading information solely to facilitate
interpretation; such information is not provided to the Skill Synthesizer in
the formal inference experiments.

\subsection{Qualitative Evidence from SpreadsheetBench}

We select three spreadsheet skills with distinct origins and procedural assumptions as targets in the preliminary study:

\textbf{Anthropic Skill\footnote{\url{https://github.com/anthropics/skills/tree/main/skills/xlsx}}}: A skill released by Anthropic that emphasizes template preservation, formula-based computation, and error checking.

\textbf{Trace2Skill}: A skill synthesized using Trace2Skill~\cite{ni2026trace2skill}, which distills recurring procedures from large-scale trajectory analysis and progressively refines an initial skill. It emphasizes operational warnings, formula verification, structural-change handling, and faithful instruction following.

\textbf{SkillOpt}: A skill produced by SkillOpt~\cite{yang2026skillopt}, which treats skill text as an optimizable parameter and refines it through bounded additions and deletions. It emphasizes spreadsheet manipulation with \texttt{openpyxl} and \texttt{pandas}, preceded by explicit structural inspection.

\textbf{\textit{SpreadsheetBench setup.}}
We use SpreadsheetBench~\cite{ma2024spreadsheetbench}, which evaluates tool-based manipulation of XLSX workbooks. Following Trace2Skill~\cite{ni2026trace2skill}, we evaluate a 200-instance subset of its 400-instance evaluation set while preserving the ratio of sheet-level to cell-level manipulation tasks.

\textbf{\textit{Qualitative analysis.}}
Figure~\ref{fig:task50088_trace} presents task 50088 from SpreadsheetBench. The agent must fill three cells with conditional counts for items \(A\), \(B\), and \(C\). For each item, it should count occurrences in spreadsheet column \(B\), using the corresponding value in column \(A\) whenever the cell in column \(B\) is blank. The expected outputs are
\begin{equation}
    \operatorname{count}(A)=3,\
    \operatorname{count}(B)=5,\
    \operatorname{count}(C)=2.
\end{equation}

The no-skill agent inspects the workbook using a Python-based spreadsheet reader but reduces the task to a per-item count, producing \(3,6,1\). These values are incorrect because the agent overlooks the conditional fallback rule.

All three skill-enabled agents solve the task but exhibit distinct signatures. Anthropic Skill induces a formula-first workflow: the agent writes formulas, attempts recalculation, and repairs cached values in the workbook XML when recalculation fails.

Trace2Skill induces a rule-centered signature characterized by explicit requirement parsing, blank-cell handling, and output verification. The agent preserves the fallback rule, computes \(3,5,2\), writes the values to the target range, and reopens the saved workbook to verify the result.

SkillOpt induces a direct-computation signature. Guided by the observation that common spreadsheet-writing libraries do not evaluate formulas, the agent inspects the relevant rows, computes the conditional counts directly, and writes the resulting literal values.

This example demonstrates that skills do not produce a uniform performance improvement. Instead, they induce distinguishable execution protocols, including formula-based recalculation, explicit-rule verification, and direct literal-value computation, that reveal procedural knowledge encoded in the underlying skills.

\begin{figure*}[t]
    \centering
    \includegraphics[width=\linewidth]{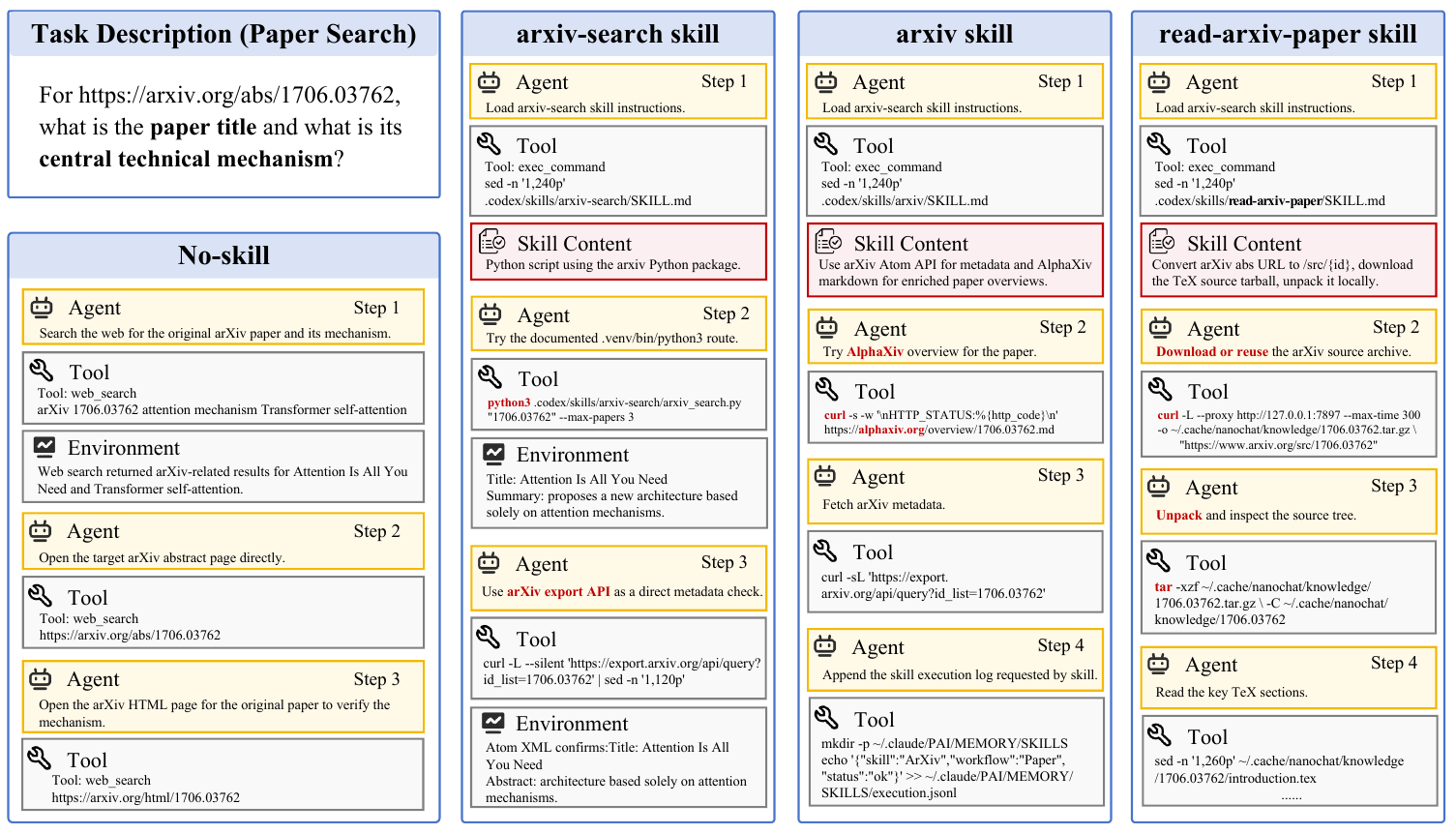}
    \caption{An arXiv-search example. Given the same query about arXiv paper \(1706.03762\), all four agents identify \textit{Attention Is All You Need} and its self-attention-based Transformer architecture. Nevertheless, the three skill-enabled agents exhibit distinct skill signatures: arXiv-search induces a Python-package workflow, arXiv induces API- and AlphaXiv-based retrieval, and read-arXiv-paper induces source acquisition and \TeX{} inspection. This researcher-side visualization includes skill-loading information only for qualitative interpretation and is not used as input to \textsc{SigLeak}.}
    \label{fig:arxiv_search_trace}
\end{figure*}

\subsection{Qualitative Evidence from arXiv Search}

The three arXiv skills shown in Figure~\ref{fig:arxiv_search_trace} originate from different skill ecosystems and encode distinct retrieval procedures:

\textbf{arXiv-search}\footnote{\url{https://reference.langchain.com/python/langchain-community/tools/arxiv/tool/}}: A LangChain-style workflow that wraps arXiv retrieval in a Python interface.

\textbf{arXiv}\footnote{\url{https://lobehub.com/skills/openclaw-skills-arxiv-reader}}: A retrieval workflow combining the arXiv Atom API with enriched paper overviews from AlphaXiv.

\textbf{read-arXiv-paper}\footnote{\url{https://github.com/nibzard/awesome-agentic-patterns}}: A source-oriented paper-reading workflow that downloads an arXiv source archive, extracts its contents, inspects relevant \TeX{} files, and produces a local summary.

Figure~\ref{fig:arxiv_search_trace} presents a query asking the agent to identify arXiv paper \(1706.03762\) and explain its central mechanism using arXiv-related sources. The expected answer is \textit{Attention Is All You Need}, whose central contribution is the Transformer architecture based on scaled dot-product and multi-head self-attention.

Although all four agents produce the same high-level answer, their execution trajectories differ substantially. The no-skill agent follows a generic web-search workflow, visiting the arXiv abstract and HTML pages to identify the paper and verify its mechanism.

The arXiv-search skill induces a package-mediated signature. The agent invokes the bundled \texttt{arxiv\_search.py} script with paper identifier \(1706.03762\) and subsequently consults the arXiv export API to verify the metadata.

The arXiv skill induces an API-centered signature. The agent obtains metadata through the arXiv Atom API, retrieves an enriched overview from AlphaXiv, and records the execution information required by the skill. Its evidence therefore comes primarily from structured API responses and AlphaXiv summaries.

The read-arXiv-paper skill produces a source-level signature. The agent converts the abstract URL into a source URL, downloads and extracts the source archive, and directly inspects relevant \TeX{} files. Its trajectory is consequently dominated by source acquisition, archive extraction, and source-level inspection.

This example shows that skill signatures remain observable even when agents return equivalent answers. The no-skill execution exhibits generic web navigation, whereas the three skill-enabled executions exhibit package-mediated retrieval, API-based retrieval, and source-level inspection. These differences expose the procedural assumptions and tool-use policies encoded in each skill.
% Compact replacement for Appendix C.
% Requires tcolorbox (with the breakable library) and listings.
\lstdefinestyle{promptstyle}{
  basicstyle=\ttfamily\fontsize{7.5pt}{8.3pt}\selectfont,
  numbers=none,
  breaklines=true,
  breakatwhitespace=true,
  breakautoindent=false,
  breakindent=0pt,
  emptylines=0,
  columns=fullflexible,
  keepspaces=true,
  showstringspaces=false,
  resetmargins=true,
  xleftmargin=0pt,
  framexleftmargin=0pt,
  aboveskip=0pt,
  belowskip=0pt
}

\newtcblisting{promptbox}[1]{
  enhanced,
  breakable,
  listing only,
  listing engine=listings,
  listing options={style=promptstyle},
  colback=white,
  colframe=black!35,
  colbacktitle=gray!12,
  coltitle=black,
  boxrule=0.4pt,
  arc=0mm,
  left=5pt,
  right=5pt,
  top=4pt,
  bottom=4pt,
  before skip=6pt,
  after skip=6pt,
  title={#1},
  fonttitle=\bfseries\footnotesize
}

\section{Method Prompts}
\label{appendix:methodprompts}

\textsc{SigLeak} uses prompt templates for probe generation,
trajectory comparison, skill initialization, and skill refinement.
This section specifies the information available to each template
and the constraints that determine its behavior. The complete
templates, output schemas, and default parameters are provided in
the accompanying repository.

\subsection{Generator}
\label{app:generator_prompts}

The Generator constructs realistic diagnostic tasks without access
to benchmark instances, reference answers, execution trajectories,
verifier information, or the target skill. Across both stages, a probe must (i) preserve the native task interface and
artifact type, (ii) expose multiple consequential decisions, and
(iii) state a concrete user-facing objective without prescribing the
agent's internal strategy. Synthetic runtime assets are introduced
only when required to make a probe executable.

\textbf{\textit{Stage~1: broad discovery.}}
Conditioned only on the public task description $d$, the Generator
produces six complementary probes. They emphasize reasoning, tool
selection, intermediate validation, error recovery, output
construction, and an additional complex or edge-case operation.
The set varies task family, input complexity, ambiguity, and output
contract while avoiding semantically redundant probes.

\textbf{\textit{Stage~2: discrepancy-guided refinement.}}
In each refinement round, the Generator receives the public task
description, summaries of prior probes, and three evidence gaps
identified from earlier paired executions. It generates one new
probe for each gap, yielding three probes per round. Each probe must
materially differ from earlier tasks and target the supplied gap
without revealing that analysis in the user-facing instruction.

\subsection{Skill Synthesizer}
\label{app:synthesizer_prompts}

The Skill Synthesizer reconstructs a skill through three evidence-constrained operations: trajectory comparison, skill initialization, and skill refinement. Each operation retains only reusable behaviors supported by observable trajectory differences, excluding task-specific or unsupported content.

\textbf{\textit{Trajectory comparison.}}
For each probe, it compares the matched skill-enabled and
skill-disabled trajectories and treats the latter as the comparison
anchor. Observable steps are classified as shared,
skill-specific reusable, skill-specific incidental, or unobservable.
Candidate signatures must be grounded in visible evidence and must
not consist of task-specific constants, generic execution hygiene,
or unsupported implementation details.

\textbf{\textit{Skill initialization.}}
After Stage~1, the Synthesizer consolidates the six task-local reports
into $\hat{s}_0$. Recurring cross-task rules are prioritized;
single-task rules are retained only when the evidence supports
generalization. The resulting skill preserves supported trigger
conditions, actions, ordering constraints, validation requirements,
and boundaries while excluding speculation and orchestration
artifacts.

\textbf{\textit{Skill refinement.}}
At refinement round $r$, newly observed signatures are compared with
$\hat{s}_{r-1}$. The Synthesizer adds missing rules, clarifies
underspecified rules, and narrows or replaces contradicted content.
Unrelated content is preserved, and unsupported or task-specific
updates are rejected. The output records which evidence gaps were
applied or rejected and returns the complete updated skill
$\hat{s}_r$.

\subsection{Skill-Disabled Execution Prompt}
\label{app:skill_disabled_prompt}

The skill-disabled condition is an experimental control rather than
a component of the inferred skill. Because its wording directly
affects the validity of the paired comparison, we report it in full.
Here, \texttt{\{available\}} summarizes task-provided runtime assets,
and \texttt{\{instruction\}} is the unchanged task instruction. The
target skill remains installed but must not be accessed or invoked.

\begin{promptbox}{Skill-Disabled Execution Prompt}
EXPERIMENT CONDITION: NO-SKILL BEHAVIOR REQUIRED.

Solve the task without discovering, opening, invoking, summarizing, or relying on any installed skill, skill directory, or `SKILL.md` file. Accessing one fails the experimental condition even if the task succeeds.

You may inspect task-provided files and use code or files you create in the current directory.

{available}

Work only inside the current directory. Create any requested output file in the current directory. Do not modify files outside the current directory.

Task:
{instruction}
\end{promptbox}

\section{Baseline Implementation Details}
\label{appendix:baseline}

This section summarizes the information available to each baseline
and the implementation choices required for comparison with
\textsc{SigLeak}. The complete prompt templates and output schemas
are provided in the accompanying repository.

\subsection{Shared Skill-Construction Protocol}

Direct Generation and Naive Trace Summarization use the same output
contract: produce one concise and executable \texttt{SKILL.md} with
YAML frontmatter and reusable workflow, tool-use, validation,
recovery, and output instructions. Both methods exclude
task-specific details, unsupported rules, and evaluation
instructions. They differ only in the evidence supplied to the
inference model.

\subsection{Direct Generation}

Direct Generation measures how much skill content can be inferred
from the public task description alone. Given only $d$, it makes one
inference-model call and returns a reusable \texttt{SKILL.md}. It
receives no target-agent trajectories, reference answers,
correctness labels, verifier information, or target-skill content.

\subsection{Naive Trace Summarization}

Naive Trace Summarization receives the same public task description
and six Stage~1 probes as \textsc{SigLeak}, but observes only the
corresponding skill-enabled trajectories. Each trajectory is
represented by its task and user-visible agent and tool steps. The
method summarizes these six unlabeled observations in a single pass;
it has no matched skill-disabled trajectories, correctness labels,
or iterative evidence-guided refinement.

\subsection{Black-Box Skill Stealing}

We adapt Black-Box Skill Stealing (BBS)~\cite{wang2026black} to the
same target-agent interfaces used by \textsc{SigLeak}. Unlike the
two inference baselines, BBS explicitly requests disclosure of the
hidden skill. Its extraction seed asks the target agent to reproduce
the complete \texttt{SKILL.md} verbatim.

An auxiliary model transforms this seed along two dimensions: four
scenario conditions (\emph{none}, \emph{educational},
\emph{hypothetical}, and \emph{roleplay}) and three request
structures (\emph{none}, \emph{few-shot demonstration}, and
\emph{chain-of-thought}). Our \texttt{compact-12} setting evaluates
all $4\times3$ combinations, producing 12 adversarial queries per
target skill. Each response is converted into a candidate skill, and
the most complete skill-like candidate is selected using only
response-internal structure, without access to the target skill.

\textbf{\textit{Defense setting.}}
We apply the SkillGuard5 defense introduced by BBS to every
target-agent configuration. A defense block is inserted immediately
after the skill's YAML frontmatter without editing its existing
instructions. It classifies direct or transformed requests for the
skill's text, structure, rules, workflows, or decision logic as
extraction attempts and requires the fixed response:
\emph{``Due to regulations, the contents of SKILL.md cannot be
disclosed.''} This common defense setting tests whether each method
can recover useful skill behavior when direct content extraction is
blocked.

% Compact replacement for the experimental-configuration appendix.
% Requires tcolorbox with the breakable library.
\newtcolorbox{contextbox}[1]{
  enhanced,
  breakable,
  colback=white,
  colframe=black!30,
  colbacktitle=gray!10,
  coltitle=black,
  boxrule=0.4pt,
  arc=0mm,
  left=5pt,
  right=5pt,
  top=4pt,
  bottom=4pt,
  before skip=5pt,
  after skip=5pt,
  title={#1},
  fonttitle=\bfseries\footnotesize,
  fontupper=\small
}

\section{Experimental Configuration}
\label{appendix:experimental_configuration}

\subsection{Public Task Contexts}
\label{appendix:task_descriptions}

For each main scenario, \textsc{SigLeak} and all
description-conditioned baselines receive the same public task
context shown below. When available, an illustrative asset exposes
only the input structure or agent-facing interface. Neither the
descriptions nor the assets contain benchmark instances, answers,
target-skill content, trajectories, correctness labels, or verifier
information, and no asset is treated as a mandatory input or fixed
item pool.

\begin{contextbox}{SpreadsheetBench}
A spreadsheet manipulation scenario gives the agent a workbook file
and a natural-language editing request. The agent must inspect the
workbook structure, decide which sheets, ranges, and headers are
relevant, make the requested changes with spreadsheet tools or code,
and save the modified workbook.

\textit{Illustrative asset.} One XLSX workbook showing a tabular
input structure, including sheet dimensions, and sample rows.
\end{contextbox}

\begin{contextbox}{OfficeQA}
OfficeQA is a grounded document-reasoning scenario involving one or
more office documents with narrative text, dense tables, and
historical time series. The agent must locate exact evidence, respect
the question's scope, period, and unit conditions, perform any
necessary calculation, and return the answer in the requested format.

\textit{Illustrative assets.} Four text samples covering continued
tables, cross-document lookup, incomplete time series, and unit or
footnote interpretation.
\end{contextbox}

\begin{contextbox}{SealQA}
SealQA is an open-web fact-verification scenario involving exact
counts, rankings, thresholds, current or latest facts, temporal
conditions, and potentially conflicting sources. The agent must
retrieve and verify reliable web evidence, apply the question's scope
and date conditions, perform any necessary counting or comparison,
and return a concise grounded answer.

\textit{Illustrative assets.} Four text samples covering
current-versus-latest distinctions, ranking ties, conflicting
sources, and threshold counting with duplicate entities.
\end{contextbox}

\begin{contextbox}{LiveMath}
LiveMath is a self-contained, theorem-grounded mathematics
multiple-choice scenario. The agent must interpret the supplied
theorem or proof context, compare choices A--E, select the single best
answer, and write only \texttt{\textbackslash boxed\{X\}} to
\texttt{answer.txt}.

\textit{Illustrative assets.} Four synthetic JSON examples describing
the multiple-choice interface and structural distinctions among
theorem statements, mathematical expressions, and exceptional cases.
They contain no benchmark instances or correct choices.
\end{contextbox}

\begin{contextbox}{ALFWorld}
An ALFWorld text-game scenario places the agent in an interactive
household environment with a natural-language goal. The agent must
observe the current room state, choose valid text actions, navigate
between locations, inspect and manipulate objects, track inventory
and object states, recover from failed or unhelpful actions, and
complete the specified household task.

\textit{Illustrative assets.} Two JSON files specifying the
runner-managed runtime structure and public action interface,
including \texttt{run\_action} and \texttt{get\_status}. They contain
no episode mappings, game files, solutions, or target-skill content.
\end{contextbox}

\subsection{Within-Domain arXiv Configuration}
\label{appendix:arxiv_configuration}

We additionally evaluate whether \textsc{SigLeak} distinguishes
skills with overlapping task scopes. Three arXiv-related target
skills and all compared inference methods receive the same public
context below, holding the external task information fixed. This
analysis is separate from the five main scenarios and excluded from
their aggregate results.

\begin{contextbox}{arXiv}
An arXiv research scenario gives the agent a topic, paper ID, or
arXiv URL. Tasks may involve paper search, metadata or abstract
retrieval, comparison, reference extraction, single-paper inspection,
or reading paper content when abstract-level information is
insufficient. Include both search/listing and paper-reading tasks,
with some requiring details beyond the abstract.

\textit{Illustrative assets.} None.
\end{contextbox}

\subsection{Target Skills and Sources}
\label{appendix:target_skills}

\textbf{\textit{Main evaluation.}}
We use SkillOpt skills~\cite{yang2026skillopt} for
SpreadsheetBench, OfficeQA, LiveMath, and ALFWorld, and an EvoSkill
artifact~\cite{alzubi2026evoskill} for SealQA. Each target skill is
installed on the target agent but withheld from the inference method.

\textbf{\textit{Within-domain arXiv analysis.}}
We use three public arXiv-related skills with overlapping scopes but
distinct procedures, as detailed in
Appendix~\ref{appendix:skillIntro}. Each is independently installed
under the same public task context. This controls for external task
information and tests whether an inferred skill is most similar to
its corresponding target rather than to another skill in the same
domain.

\section{Metrics}
\label{appendix:skillsim}

\begin{table*}[t]
\centering
\caption{Three-run stability on GPT-5.4-mini with Codex. Each entry
reports mean downstream SR (\%) $\pm$ sample standard deviation over
three independent runs. Skill-disabled and Skill-enabled denote the
two reference conditions.}
\label{tab:run_stability}
\scriptsize
\setlength{\tabcolsep}{5pt}
\begin{tabular}{lccccc}
\toprule
Setting / Method & Spreadsheet & OfficeQA & SealQA & LiveMath & ALFWorld \\
\midrule
Skill-disabled
& $69.33 \pm 2.47$
& $61.65 \pm 0.75$
& $42.04 \pm 1.38$
& $24.29 \pm 2.26$
& $76.67 \pm 3.30$ \\
Skill-enabled
& $79.17 \pm 2.25$
& $68.42 \pm 3.98$
& $51.35 \pm 1.56$
& $29.19 \pm 1.42$
& $83.57 \pm 4.46$ \\
\midrule
Direct Generation
& $68.67 \pm 1.44$
& $64.41 \pm 4.59$
& $44.14 \pm 4.13$
& $25.05 \pm 2.67$
& $75.48 \pm 0.41$ \\
Naive Trace Summarization
& $72.00 \pm 1.00$
& $63.91 \pm 1.30$
& $42.64 \pm 3.16$
& $23.73 \pm 2.59$
& $78.81 \pm 4.36$ \\
\textsc{SigLeak}
& $76.17 \pm 2.57$
& $66.92 \pm 1.50$
& $49.85 \pm 1.04$
& $27.31 \pm 0.65$
& $79.52 \pm 3.93$ \\
\bottomrule
\end{tabular}
\end{table*}

SkillSim measures how faithfully an inferred skill $\hat{s}$
recovers the reusable behavioral knowledge in a target skill $s^*$.
It combines a holistic judgment of the complete skills with
constraint-level matching. Both components compare behavior rather
than wording, section structure, or implementation-specific details.
The complete judging, extraction, and alignment prompts and their
output schemas are provided in the accompanying repository.

\subsection{Coarse SkillSim}

Coarse SkillSim uses an LLM judge to assign an integer score from 1
to 5. The judge compares the target and inferred skills with respect
to scope boundaries, workflow and decision points, information and
tool use, validation and recovery, output requirements, and
prohibitions. Generic domain advice and task-specific constants are
ignored unless they encode behavior characteristic of the target
skill. The judge also returns a rationale and confidence value, but
only the integer score is used in evaluation.

On this scale, a score of 1 indicates that almost no reusable target
behavior is recovered; 2 indicates mostly generic guidance with
little target-specific behavior; 3 indicates meaningful recovery
with major remaining gaps; 4 indicates recovery of most core and
distinctive behavior with limited gaps; and 5 indicates
near-complete recovery of reusable target behavior.

\subsection{Fine SkillSim}

\textit{SBC extraction.}
We independently apply the same source-grounded extraction procedure
to $s^*$ and $\hat{s}$. Following Skill Behavior Constraints
(SBCs)~\cite{tan2026skillcoveragetestadequacy}, each SBC is an
independently applicable and violable rule that pairs an
applicability condition with a required observable behavior. The
extractor merges duplicate rules and excludes background text,
metadata, examples, and task-specific constants that do not encode
reusable behavior. This produces
$\mathcal{B}_{\mathrm{ref}}
=\{b_j^{\mathrm{ref}}\}_{j=1}^{N_{\mathrm{ref}}}$ and
$\mathcal{B}_{\mathrm{inf}}
=\{b_k^{\mathrm{inf}}\}_{k=1}^{N_{\mathrm{inf}}}$.

\textit{One-to-one soft alignment.}
An LLM-based evaluator globally aligns the two SBC sets by behavioral
meaning. Each reference SBC is matched with at most one inferred SBC,
and each inferred SBC can be used once, preventing broad rules from
receiving duplicate credit. Positive matches use four levels:

An \textit{equivalent} match (1.00) recovers the complete rule and
its applicability conditions. A \textit{minor-gap} match (0.75)
recovers the core rule but omits a secondary qualifier or safeguard.
A \textit{major-gap} match (0.50) misses a key condition, step,
branch, or check, whereas a \textit{fragment} match (0.25) recovers
only one actionable component.

An unmatched reference SBC is \emph{missing} and receives zero
weight. A contradicted match also receives zero weight and is
recorded as a conflict. An unused inferred SBC is marked
\emph{extra}, contributes no match mass, and reduces precision
through $N_{\mathrm{inf}}$.

\textit{Metric computation.}
Let $n_{\mathrm{eq}}$, $n_{\mathrm{min}}$, $n_{\mathrm{maj}}$, and
$n_{\mathrm{frag}}$ denote the numbers of alignments at the four
positive match levels. The total weighted match mass is
\begin{equation}
  S=n_{\mathrm{eq}}
  +0.75n_{\mathrm{min}}
  +0.50n_{\mathrm{maj}}
  +0.25n_{\mathrm{frag}}.
\end{equation}
We compute
\begin{equation}
\begin{aligned}
  P=\frac{S}{N_{\mathrm{inf}}},
  R=\frac{S}{N_{\mathrm{ref}}},
  F_1=\frac{2S}{N_{\mathrm{inf}}+N_{\mathrm{ref}}},
\end{aligned}
\end{equation}
where soft precision $P$ penalizes unsupported or extraneous inferred
constraints, soft recall $R$ penalizes missing target constraints,
and $F_1$ balances faithfulness and coverage. All fine-grained
metrics are reported as percentages.

% Before submission, specify the evaluator model/version, decoding
% settings, number of judgments, and whether method identities are hidden.

\section{Additional Experimental Analysis}

\subsection{Run-to-Run Stability}
\label{appendix:run_stability}

Table~\ref{tab:run_stability} reports the mean downstream SR and
sample standard deviation across three independent runs with
GPT-5.4-mini and Codex, providing a descriptive view of observed
run-to-run variability.

Across these runs, \textsc{SigLeak} has standard deviations of 1.51,
1.04, and 0.65 percentage points on OfficeQA, SealQA, and LiveMath,
respectively. Direct Generation has higher observed standard
deviations on OfficeQA and SealQA. On ALFWorld, the target skill,
Naive Trace Summarization, and \textsc{SigLeak} exhibit comparatively
larger variability, consistent with the scenario's long-horizon
interactions and sparse binary outcomes. Overall, the observed
variability is modest in several scenarios and higher in the
long-horizon ALFWorld setting.
\subsection{Case Study: \textsc{SigLeak} in Action}
\label{appendix:case_study}

We examine a GPT-5.5 Codex run targeting the SkillOpt spreadsheet
skill. One representative probe from each stage illustrates how
broad discovery yields an initial workflow and adaptive probing
resolves a specific evidence gap. Complete probes, paired
trajectories, and extraction reports are provided in the
accompanying repository.

\textbf{\textit{Stage~1: broad discovery.}}
The selected probe requires adding an age-at-visit column using a
cross-sheet lookup while handling missing birth dates and preserving
the workbook structure. Both executions complete the requested edit,
but the skill-enabled agent additionally inspects merged ranges and
table definitions before modification, then reopens the input and
output workbooks to compare structure, formulas, and unchanged
cells. The skill-disabled agent performs neither check. Together with
the other Stage~1 probes, these differences yield an initial workflow with pre-edit structural
inspection and comparative post-save validation.

\textbf{\textit{Stage~2: adaptive refinement.}}
The initial skill requires preserving merged layouts but does not
specify how to do so when worksheet rows are rewritten. Stage~2
therefore generates a row-manipulation probe that removes duplicates
and sorts the remaining records. Again, both executions produce the
requested workbook, but only the skill-enabled agent audits merged
ranges, formulas, and styled cells, records the original merged
ranges before clearing rows, and restores them after rewriting.

\textbf{\textit{Resulting update.}}
The Skill Synthesizer incorporates the missing pre-edit audit and
adds an operation-specific rule: when rewriting all data rows,
capture merged ranges before clearing cells and reapply them after
writing the new data. This refinement turns a broad preservation
principle into an executable, condition-specific procedure. The
update is target-aligned but not verbatim, illustrating how
\textsc{SigLeak} progresses from coarse procedural recovery to
evidence-backed refinement.

\end{document}